\documentclass[11pt,a4paper]{article}
\usepackage[hyperref]{emnlp2020}
\usepackage{times,graphicx,multirow,amsmath,subfigure,caption,amssymb,array,booktabs,xcolor}
\usepackage{latexsym}

\usepackage{microtype}

\aclfinalcopy 

\title{A Probabilistic End-To-End Task-Oriented Dialog Model with Latent Belief States towards Semi-Supervised Learning}

\author{Yichi Zhang$^{1}$, Zhijian Ou$^{1}$\thanks{~~Corresponding author.}~, Min Hu$^2$, Junlan Feng$^{2}$  \\
	$^1$ Speech Processing and Machine Intelligence Lab, Tsinghua University, Beijing, China \\
	$^2$ China Mobile Research Institute, Beijing, China \\
	{\tt zhangyic17@tsinghua.org.cn}, {\tt ozj@tsinghua.edu.cn}}

\date{}

\newcommand{\modelname}{LABES}

\begin{document}
	\maketitle
	\begin{abstract}
		Structured belief states are crucial for user goal tracking and database query in task-oriented dialog systems. However, training belief trackers often requires expensive turn-level annotations of every user utterance.
		In this paper we aim at alleviating the reliance on belief state labels in building end-to-end dialog systems, by leveraging unlabeled dialog data towards semi-supervised learning.
		We propose a probabilistic dialog model, called the LAtent BElief State (\modelname{}) model, where belief states are represented as discrete latent variables and jointly modeled with system responses given user inputs.
		Such latent variable modeling enables us to develop semi-supervised learning under the principled variational learning framework.
		Furthermore, we introduce LABES-S2S, which is a copy-augmented Seq2Seq model instantiation of LABES\footnote{Code available at https://github.com/thu-spmi/LABES}.
		In supervised experiments, LABES-S2S obtains strong results on three benchmark datasets of different scales. In utilizing unlabeled dialog data, semi-supervised LABES-S2S significantly outperforms both supervised-only and semi-supervised baselines.
		Remarkably, we can reduce the annotation demands to 50\% without performance loss on MultiWOZ. 
	\end{abstract}
	
	\section{Introduction}
\begin{figure}[t]
		\centering
		\includegraphics[width=1.0\columnwidth]{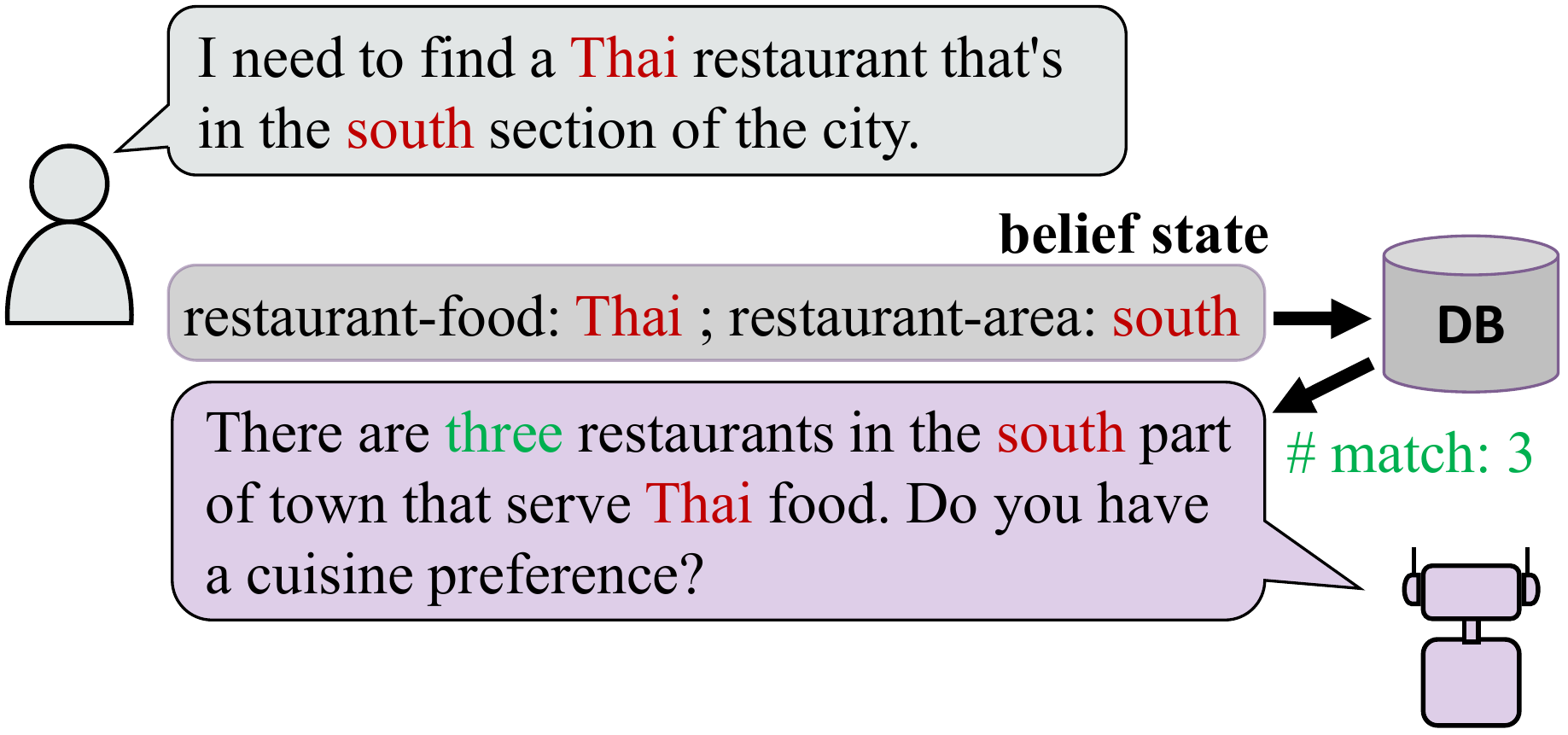}
		\caption{The cues for inferring belief states from user inputs and system responses. The system response reveals the belief state either directly in the form of word repetition (red), or indirectly in the form of the database query result (green) determined by the belief state. 
		}
		\label{dial_example}
	\end{figure}

	Belief tracking (also known as dialog state tracking) is an important component in task-oriented dialog systems. The system tracks user goals through multiple dialog turns, i.e. infers structured \textit{belief states} expressed in terms of slots and values (e.g. in Figure \ref{dial_example}), to query an external database \cite{henderson2014second}.
	Different belief tracking models have been proposed in recent years, either trained independently \cite{mrkvsic2017neural,ren2018towards,wu2019transferable} or within end-to-end (E2E) trainable dialog systems \cite{wen2017latent,wen2017a,liu2017end,lei2018sequicity,fsdm,liang2020moss,zhang2020task}.
	
	Existing belief trackers mainly depend on supervised learning with human annotations of belief states for every user utterance. However, collecting these turn-level annotations is labor-intensive and time-consuming, and often requires domain knowledge to identify slots correctly. Building E2E trainable dialog systems, called E2E dialog systems for short, even further magnifies the demand for increased amounts of labeled data \cite{gao2020paraphrase,zhang2020task}. 
	
	Notably, there are often easily-available unlabeled dialog data such as between customers and trained human agents accumulated in real-world customer services.
	In this paper, we are interested in reducing the reliance on belief state annotations in building E2E task-oriented dialog systems, by leveraging unlabeled dialog data towards semi-supervised learning.
	Intuitively, the dialog data, even unlabeled, can be used to enhance the performance of belief tracking and thus benefit the whole dialog system, because there are cues from user inputs and system responses which reveal the belief states, as shown in Figure \ref{dial_example}.
	
	
	Technically, we propose a latent variable model for task-oriented dialogs, called the \textbf{LA}tent \textbf{BE}lief \textbf{S}tate (\modelname{}) dialog model.
	The model generally consists of multiple (e.g. $\!T$) turns of user inputs $u_{1:T}$ and system responses $r_{1:T}$ which are observations, and belief states $b_{1:T}$ which are latent variables.
	Basically, \modelname{} is a conditional generative model of belief states and system responses given user inputs, i.e. $p_\theta(b_{1:T},r_{1:T}|u_{1:T})$.
	Once built, the model can be used to infer belief states and generate responses.
	More importantly, such latent variable modeling enables us to develop semi-supervised learning on a mix of labeled and unlabeled data under the principled variational learning framework \cite{kingma2013auto,sohn2015learning}.
	In this manner, we hope that the LABES model can exploit the cues for belief tracking from user inputs and system responses.
	Furthermore, we develop \modelname{}-S2S, which is a specific model instantiation of \modelname{}, employing copy-augmented Seq2Seq \cite{gu2016incorporating} based conditional distributions in implementing $p_\theta(b_{1:T},r_{1:T}|u_{1:T})$. 
	
	
	
	We show the advantage of our model compared to other E2E task-oriented dialog models, and demonstrate the effectiveness of our semi-supervised learning scheme on three benchmark task-oriented datasets: CamRest676 \cite{wen2017a}, In-Car \cite{eric2017key} and MultiWOZ \cite{budzianowski2018multiwoz} across various scales and domains. 
	In supervised experiments, \modelname{}-S2S obtains state-of-the-art results on CamRest676 and In-Car, and outperforms all the existing models which do not leverage large pretrained language models on MultiWOZ. 
	In utilizing unlabeled dialog data, semi-supervised \modelname{}-S2S significantly outperforms both supervised-only and prior semi-supervised baselines. 
	Remarkably, we can reduce the annotation requirements to 50\% without performance loss on MultiWOZ, which is equivalent to saving around 30,000 annotations.

	\section{Related Work}
	
	\label{sec:related}
	\paragraph{On use of unlabeled data for belief tracking.}
	Classic methods such as self-training \cite{rosenberg2005semi}, also known as pseudo-labeling \cite{lee2013pseudo}, has been applied to belief tracking \cite{tseng2019semi}.
	Recently, the pretraining-and-fine-tuning approach has received increasing interests \cite{heck2020trippy,peng2020soloist,hosseini2020simple}. 
	The generative model based semi-supervised learning approach, which blends unsupervised and supervised learning, has also been studied \cite{wen2017latent, sedst}.
	Notably, the two approaches are orthogonal and could be jointly used. 
	Our work belongs to the second approach, aiming to leverage unlabeled dialog data beyond of using general text corpus.
	A related work close to ours is SEDST \cite{sedst}, which also perform semi-supervised learning for belief tracking.
	Remarkably, our model is optimized under the principled variational learning framework, while SEDST is trained with an ad-hoc combination of posterior regularization and auto-encoding.
	Experimental in \S\ref{sec:semi_exp} show the superiority of our model over SEDST. 
	See Appendix \ref{app:vs} for differences in model structures between SEDST and \modelname{}-S2S. 
	
	
	\paragraph{End-to-end task-oriented dialog systems.}
	Our model belongs to the family of E2E task-oriented dialog models \cite{wen2017latent,wen2017a,li2017end,lei2018sequicity,mehri2019structured,wu2019alternating,peng2020soloist,hosseini2020simple}. We borrow some elements from the Sequicity \cite{lei2018sequicity} model, such as representing the belief state as a natural language sequence (a text span), and using copy-augmented Seq2Seq learning \cite{gu2016incorporating}. 
	But compared to Sequicity and all its follow-up works \cite{sedst,fsdm,zhang2020task,liang2020moss}, a feature in our LABES-S2S model is that the transition between belief states across turns and the dependency between system responses and belief states are well statistically modeled.
	This new design results in a completely different graphical model structure, which enables rigorous probabilistic variational learning. See Appendix \ref{app:vs} for details. 
	
	
	\paragraph{Latent variable models for dialog.}
	Latent variables have been used in dialog models. 
	For non task-oriented dialogs, latent variables are introduced to improve diversity \cite{serban2017hierarchical,zhao2017learning,gao2019a}, control language styles \cite{gao2019structuring} or incorporate knowledge \cite{kim2020sequential} in dialog generation. 
	For task-oriented dialogs, there are prior studies which use latent internal states via hidden Markov models \cite{zhai2014discovering} or variational autoencoders \cite{shi2019unsupervised} to discover the underlying dialog structures. 
	In \citet{wen2017latent} and \citet{zhao2019rethinking}, dialog acts are treated as latent variables, together with variational learning and reinforcement learning, aiming to improve response generation. 
	To the best of our knowledge, we are the first to model belief state as discrete latent variables, and propose to learn these structured representations via the variational principle. 
	
	\section{Latent Belief State Dialog Models}
	\label{sec:model}
	We first introduce \modelname{} as a general dialog modeling framework in this section. For dialog turn $t$, let $u_t$ be the user utterance, $b_t$ be the current belief state after observed $u_t$ and $r_t$ be the corresponding system response. In addition, denote $c_t$ as the dialog context or model input at turn $t$, such as $c_t \triangleq \{r_{t-1}, u_t\}$ as in this work. Note that $c_t$ can include longer dialog history depending on specific implementations. Let $d_t$ be the database query result which can be obtained through a database-lookup operation given the belief state $b_t$. 
	
	Our goal is to model the joint distribution of belief states and system responses given the user inputs, $p_\theta(b_{1:T},r_{1:T}|u_{1:T})$, where $T$ is the total number of turns and $\theta$ denotes the model parameters. In \modelname{}, we assume the joint distribution follows the directed probabilistic graphical model illustrated in Figure \ref{pgm}, which can be formulated as:
	\begin{align*}
	p_\theta(b_{1:\!T}, r_{1:\!T}|u_{1:\!T}\!) \!\!&=\!p_\theta(b_{1:\!T}|u_{1:\!T})p_\theta(r_{1:\!T}|b_{1:\!T},u_{1:\!T}\!) \\
	&=\!\!\prod_{t=1}^T\! p_\theta(b_t|b_{t-\!1}, \!c_t)p_\theta(r_t|c_t, \!b_t, \!{d}_t)
	\end{align*}
	where $b_{0}$ is an empty state. 
	Intuitively, we refer the conditional distribution $p_\theta(b_t|b_{t-\!1}, \!c_t)$ as the belief state decoder, and $p_\theta(r_t|c_t, \!b_t, \!{d}_t)$ the response decoder in the above decomposition.
	Note that the probability $p(d_{t}|b_t)$ is omitted as database result $d_t$ is deterministically obtained given $b_t$. 
	Thus the system response can be generated as a three-step process: first predict the belief state $b_t$, then use $b_t$ to query the database and obtain $d_t$, finally generate the system response $r_t$ based on all the conditions. 
	
	\subsection*{Unsupervised Learning}
\begin{figure}[t]
		\centering
		\includegraphics[width=0.85\columnwidth]{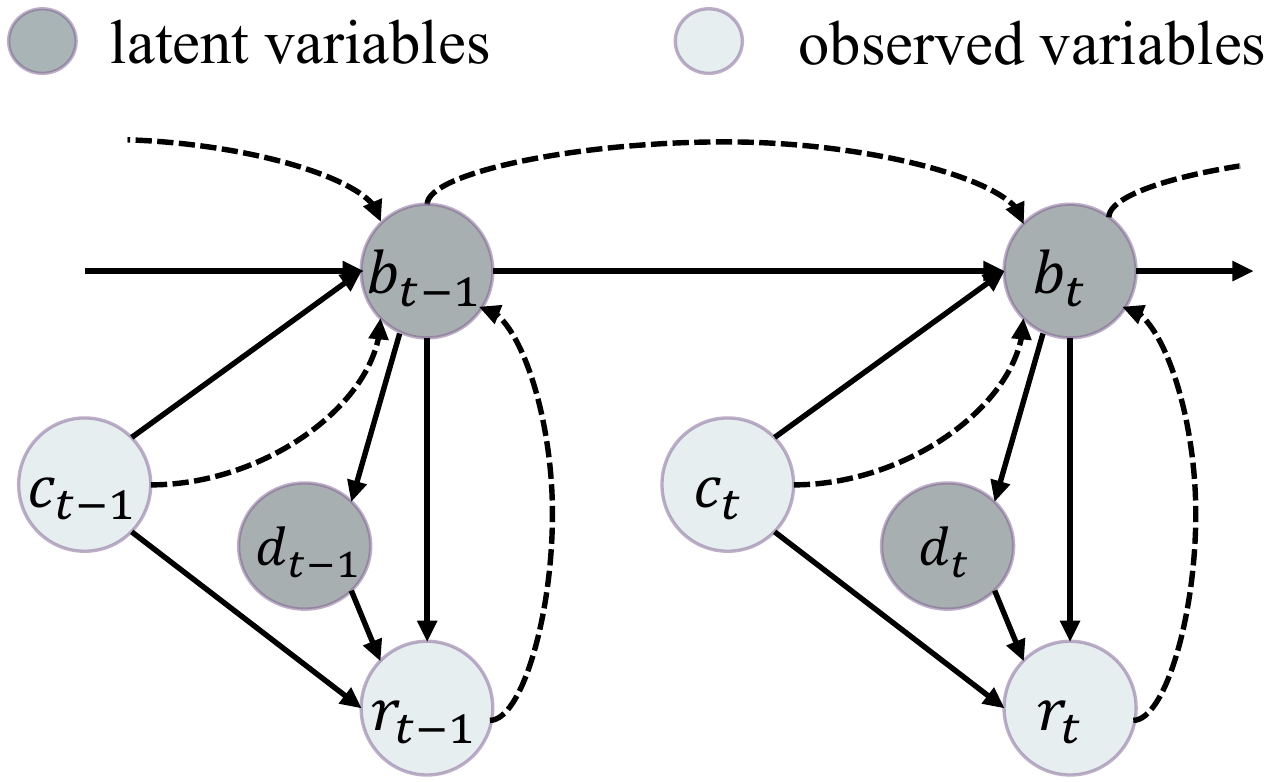}
		\caption{The probabilistic graphical model of \modelname{}. Solid arrows describe the conditional generative model $p_\theta$, and dash arrows describe the approximate posterior model $q_\phi$. Note that we set $c_t\triangleq\{r_{t-1}, u_t\}$ in our model, and omit $u_t$ from the graph for simplicity.  }
		\label{pgm}
	\end{figure}

	We introduce an inference model $q_\phi(b_t|b_{t-1}, c_t, r_t)$ (described by dash arrows in Figure \ref{pgm}) to approximate the true posterior $p_\theta(b_t|b_{t-1}, c_t, r_t)$. Then we can derive the variational evidence lower bound (ELBO) for unsupervised learning as follows:
	\begin{align*}
	\mathcal{J}_{un}
	\!=&~\mathbb{E}_{q_\phi(b_{1:T})}\bigg[\log \frac{p_\theta(b_{1:T},r_{1:T}|u_{1:T})}{q_\phi(b_{1:T}|u_{1:T},r_{1:T})} \bigg]\nonumber \\
	\!=&\sum_{t=1}^T \mathbb{E}_{q_\phi(b_{1:T})}\big[\log p_\theta(r_t|c_t,b_t,d_t)\big] \\
	\!&- \!\alpha\text{KL}\big[q_\phi(b_t|b_{t-1},c_t, r_t)\lVert p_\theta(b_t|b_{t-1},c_t)\big] \nonumber
	\end{align*}
	where 
	\begin{align*}
	q_\phi(b_{1:T})&\triangleq \prod_{t=1}^T q_\phi(b_{t}|b_{t-1}, c_{t}, r_{t})
	\end{align*}
	and $\alpha$ is a hyperparameter to control the weight of the KL term introduced by \citet{higgins2017beta}. 
	
	Optimizing $\mathcal{J}_{un}$ requires drawing posterior belief state samples $b_{1:T}\sim q_\phi(b_{1:T}|u_{1:T},r_{1:T})$ to estimate the expectations. Here we use a sequential sampling strategy similar to \citet{kim2020sequential}, where each $b_t$ sampled from $q_\phi(b_t|b_{t-1},c_t, r_t)$ at turn $t$ is used as the condition to generate the next turn's belief state $b_{t+1}$. For calculating gradients with discrete latent variables, which is non-trivial, some methods have been proposed such as using a score function estimator \cite{williams1992simple} or categorical reparameterization trick \cite{jang2016categorical}. In this paper, we employ the simple Straight-Through estimator \cite{bengio2013estimating}, where the sampled discrete token indexes are used for forward computation, and the continuous softmax probability of each token is used for backward gradient calculation. Although the Straight-Through estimator is biased, we find it works pretty well in our experiments, therefore leave the exploration of other optimization methods as future work. 
	
	\begin{figure*}[t]
		\centering
		\subfigure[Overview of \modelname{-S2S}.]
		{	\label{all}
			\includegraphics[width=0.58\linewidth]{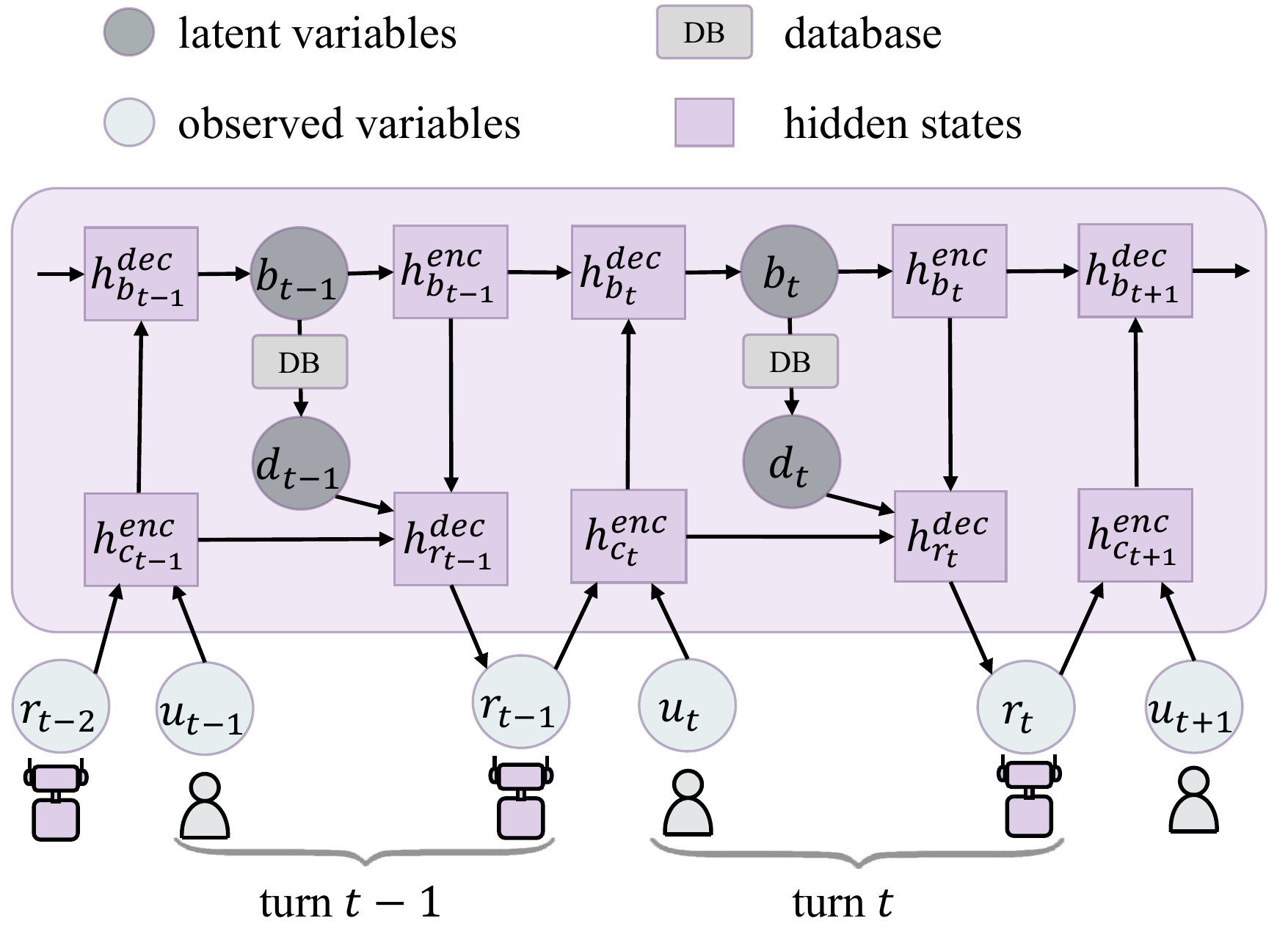} }
		\subfigure[Structure of the belief state decoder.]
		{	\label{decoder}
			\includegraphics[width=0.38\linewidth]{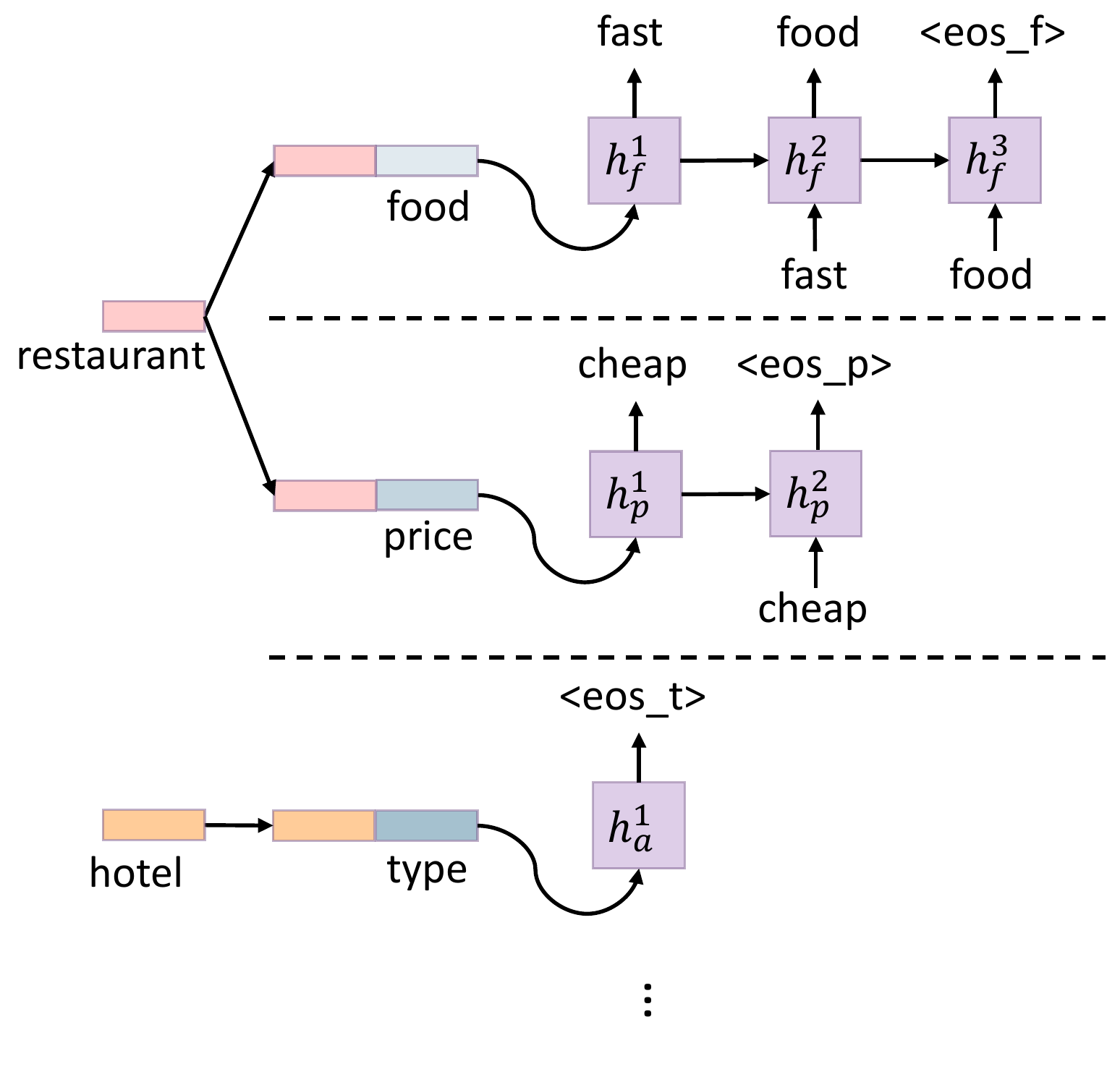} }
		\caption{ (a) shows the computational graph of \modelname{-S2S}. In (b), rectangles in different colors denote different word embeddings, and the embedding of domain names and slot names are concatenated as the initial input. Note that the same (i.e. weight-tied) decoder is shared across all slots. Decoding stops when a slot-specific end-of-sentence symbol is generated, which is possible to be the first output if the slot does not appear in the dialog. }
		\label{model}
	\end{figure*}
	
	\subsection*{Semi-Supervised Learning}
	When $b_t$ labels are available, we can easily train the generative model $p_\theta$ and inference model $q_\phi$ via supervised maximum likelihoods:
	\begin{align*}
	\mathcal{J}_{sup}
	\!&=\!\!\!\sum_{t=1}^T \!\big[\log p_\theta(b_{t}|b_{t-1}, c_t)
	\!\!+\!\log p_\theta(r_t|c_t, b_t, d_t) \\
	\!&\qquad~~+\log q_\phi(b_t|b_{t-1},c_t, r_t)\big]
	\label{eq2}
	\end{align*}
	
	When a mix of labeled and unlabeled data is available, we perform semi-supervised learning using a combination of the supervised objective $\mathcal{J}_{sup}$ and the unsupervised objective $\mathcal{J}_{un}$.
	Specifically, we first pretrain $p_\theta$ and $q_\phi$ on small-sized labeled data until convergence. Then we draw supervised and unsupervised minibatches from labeled and unlabeled data and perform stochastic gradient ascent over $\mathcal{J}_{sup}$ and $\mathcal{J}_{un}$, respectively. We use supervised pretraining first because training $q_\phi(b_t|b_{t-1},c_t, r_t)$ to correctly generate slot values and special outputs such as ``dontcare" and end-of-sentence tokens 
	as much as possible is important to improve sample efficiency in subsequent semi-supervised learning.

	\section{\modelname{}-S2S: A Copy-Augmented Seq2Seq Instantiation}
	\label{sec:impl}
	
	In the above probabilistic dialog model \modelname{}, the belief state decoder $p_\theta(b_t|b_{t-\!1}, \!c_t)$ and the response decoder $p_\theta(r_t|c_t, \!b_t, \!{d}_t)$ can be flexibly implemented.
	In this section we introduce \modelname{}-S2S as an instantiation of the general \modelname{} model based on copy-augmented Seq2Seq conditional distributions \cite{gu2016incorporating}, which is shown in Figure \ref{all} and described in the following.
	The responses are generated through two Seq2Seq processes: 1) decode the belief state given dialog context and last turn's belief state and 2) decode the system response given dialog context, the decoded belief state and database query result. 

	\subsection*{Belief State Decoder} 
	The belief state decoder is implemented via a Seq2Seq process, as shown in Figure \ref{decoder}. Inspired by \citet{fsdm}, we use a single GRU decoder to decode the value for each informable slot separately, feeding the embedding of each slot name as the initial input. In multi-domain setting, the domain name embedding is concatenated with the slot name embedding to distinguish slots with identical names in different domains \cite{wu2019transferable}. 
	
	We use two bi-directional GRUs \cite{cho2014learning} to encode the dialog context $c_t$ and previous belief state $b_{t-1}$ into a sequence of hidden vectors $h_{c_t}^{enc}$ and $h_{b_{t-1}}^{enc}$ respectively, which are the inputs to the belief state decoder. As there are multiple slots, and their values can also consist of multiple tokens, we denote the $i$-th token of slot $s$ by $b_t^{s,i}$. To decode each token $b_t^{s,i}$, we first compute an attention vector over the encoder vectors. Then the attention vector and the embedding of the last decoded token $e(b_t^{s,i-1})$ are concatenated and fed into the decoder GRU to get the decoder hidden state $h_{b_t^{s,i}}^{dec}$, denoted as $h_{s,i}^{dec}$ for simplicity:
	\begin{align*}
	a_t^{s,i} &= \mathrm{Attn}(h_{c_t}^{enc}\circ h_{b_{t-1}}^{enc}, h_{s,i}^{dec}) \\
	h_{s,i}^{dec} &= \mathrm{GRU}(a_t^{s,i}\circ e(b_t^{s,i-1}), h_{s,i-1}^{dec}) \\
	\hat{h}_{s,i}^{dec} &= \mathrm{dropout}\big(h_{s,i}^{dec}\circ e(b_t^{s,i-1})\big)
	\end{align*}
	where $\circ$ denotes vector concatenation. We use the last hidden state of the dialog context encoder as $h_{s,0}^{dec}$, and the slot name embedding as $e(b_t^{s,0})$. We reuse $e(b_t^{s,i-1})$ to form $\hat{h}_{s,i}^{dec}$ to give more emphasis on the slot name embedding and add a dropout layer to reduce overfitting. $\hat{h}_{s,i}^{dec}$ is then used to compute a generative score $\psi_{gen}$ for each token $w$ in the vocabulary $\mathcal{V}$, and a copy score $\psi_{cp}$ for words appeared in $c_t$ and $b_{t-1}$. Finally, these two scores are combined and normalized to form the final decoding probability following:
	\begin{align*}
	\psi_{gen}(b_t^{s,i}\!=w)\! &= v_w^\mathsf{T}\mathrm{W_{gen}}\hat{h}_{s,i}^{dec}, \quad w\in \mathcal{V} \\
	\psi_{cp}(b_t^{s,i}\!=x_j)\! &=  h_{x_j}^{enc\mathsf{T}}\mathrm{W_{cp}}\hat{h}_{s,i}^{dec}, \!\!\quad x_j\in c_t\cup b_{t-1} \\
	p(b_t^{s,i}\!=w)\! &= \frac{1}{Z} \bigg(e^{\psi_{gen}(w)} + \sum_{j:x_j=w} e^{\psi_{cp}(x_j)}\bigg)
	\end{align*}
	where $\mathrm{W_{gen}}$ and $\mathrm{W_{cp}}$ are trainable parameters, $v_w$ is the one-hot representation of $w$, $x_j$ is the $j$-th token in $c_t\cup b_{t-1}$ and $Z$ is the normalization term. 
	
	With copy mechanism, it is easier for the model to extract words mentioned by the user and keep the unchanged values from previous belief state. Meanwhile, the decoder can also generate tokens not appeared in input sequences, e.g. the special token ``dontcare" or end-of-sentence symbols. Since the decoding for each slot is independent with each other, all the slots can be decoded in parallel to speed up. 
	
	The posterior network $q_\phi(b_t|b_{t-1},c_t,r_t)$ is constructed through a similar process, where the only difference is that the system response $r_t$ is also encoded and used as an additional input to the decoder. Note that the posterior network is separately parameterized with $\phi$.

	\subsection*{Response Decoder}
	\label{sec:resp_dec}
	The response decoder is implemented via another Seq2Seq process.
	After obtaining the belief state $b_t$, we use it to query a database to find entities that meet user's need, e.g. Thai restaurants in the south area. The query result $d_t$ is represented as a 5-dimension one-hot vector to indicate 0, 1, 2, 3 and \textgreater3 matched results respectively. We only need the number of matched entities instead of their specific information as the input to the response decoder, because we generate delexicalized responses with placeholders for specific slot values (as shown in Table \ref{case}) to improve data efficiency \cite{wen2015semantically}. The values can be filled through simple rule-based post-processing afterwards. 
	
	Instead of directly decoding the response from the belief state decoder's hidden states \cite{lei2018sequicity}, 
	we again use the bi-directional GRU (the one used to encode $b_{t-1}$) to encode the current belief state $b_{t}$ into hidden vectors $h_{b_t}^{enc}$. Then for each token $r_{t}^i$ in the response, the decoder state $h_{r_{t,i}}^{dec}$ can be computed as follows:
	\begin{align*}
	a_t^{i} &= \mathrm{Attn}(h_{c_t}^{enc}\circ h_{b_{t}}^{enc}, h_{r_{t,i}}^{dec}) \\
	h_{r_{t,i}}^{dec} &= \mathrm{GRU}(a_t^{i}\circ e(r_t^{i-1})\circ d_t, h_{r_{t,i-1}}^{dec})\\
	\hat{h}_{r_{t,i}}^{dec} &= h_{r_{t,i}}^{dec}\circ a_t^i \circ d_t
	\end{align*}
	Note that dropout is not used for $\hat{h}_{r_{t,i}}^{dec}$, since response generation is not likely to overfit, compared to belief tracking in practice.  
	We omit the probability formulas because they are almost the same as in the belief state decoder, except for changing the copy source from $c_t\cup b_{t-1}$ to $c_t\cup b_{t}$. 
	
	\section{Experimental Settings}
	\subsection{Datasets}
	We evaluate the proposed model on three benchmark task-oriented dialog datasets: the Cambridge Restaurant (CamRest676) \cite{wen2017a}, Stanford In-Car Assistant (In-Car) \cite{eric2017key} and MultiWOZ \cite{budzianowski2018multiwoz}, with 676/3031/10438 dialogs respectively. In particular, MultiWOZ is one of the most challenging dataset up-to-date given its multi-domain setting, complex ontology and diverse language styles. As there are some belief state annotation errors in MultiWOZ, we use the corrected version MultiWOZ 2.1 \cite{eric2019multiwoz} in our experiments. See Appendix \ref{app:data} for more detailed introductions and statistics.
	
	\subsection{Evaluation Metrics}
	We evaluate the model performance under the end-to-end setting, i.e. the model needs to first predict belief states and then generate response based on its own belief predictions. For evaluating belief tracking performance, we use the commonly used \texttt{joint goal accuracy}, which is the proportion of dialog turns where all slot values are correctly predicted. For evaluating response generation, we use \texttt{BLEU} \cite{papineni2002bleu} to measure the general language quality. The response quality towards task completion is measured by dataset-specific metrics to facilitate comparison with prior works. For CamRest676 and In-Car, we use \texttt{Match} and \texttt{SuccF1} following \citet{lei2018sequicity}. 
	For MultiWOZ, we use \texttt{Inform} and \texttt{Success} as in \citet{budzianowski2018multiwoz}, and also a combined score computed through (\texttt{Inform}$+$\texttt{Success})$\times 0.5+$\texttt{BLEU} as the overall response quality suggested by \citet{mehri2019structured}. 
	
	\subsection{Baselines}
	In our experiments, we compare our model to various Dialog State Tracking (DST) and End-to-End (E2E) baseline models. 
	Recently, large-scale pretrained language models (LM) such as BERT \cite{devlin2019bert} and GPT-2 \cite{radford2019gpt2} are used to improve the performance of dialog models, however in the cost of tens-fold larger model sizes and computations. We distinguish them from light-weighted models trained from scratch in our comparison. 
	
	\paragraph{Independent DST Models:}
	For CamRest676, we compare to StateNet \cite{ren2018towards} and TripPy \cite{heck2020trippy}, which are the SOTA model without/with BERT respectively. For MultiWOZ, we compare to BERT-free models TRADE \cite{wu2019transferable}, NADST \cite{le2020nadst} and CSFN-DST \cite{zhu2020efficient}, and BERT-based models including TripPy, the BERT version of CSFN and DST-Picklist \cite{zhang2019find}. 
	
	\paragraph{E2E Models:} E2E models can be divided into three sub-categories. The TSCP \cite{lei2018sequicity}, SEDST \cite{sedst}, FSDM \cite{fsdm}, MOSS \cite{liang2020moss} and DAMD \cite{zhang2020task} are based on the copy-augmented Seq2Seq learning framework proposed by \citet{lei2018sequicity}. LIDM \cite{wen2017latent}, SFN \cite{mehri2019structured} and UniConv \cite{le2020uniconv} are modular designed, connected through neural states and trained end-to-end. SimpleTOD \cite{hosseini2020simple} and SOLOLIST \cite{peng2020soloist} are two recent models, which both use a single auto-regressive language model, initialized from GPT-2, to build the entire system. 
	
	\paragraph{Semi-Supervised Methods:}
	\label{sec:baselines}
	First, we compare with SEDST \cite{sedst} for semi-supervised belief tracking performance. SEDST is also a E2E dialog model based on copy-augmented Seq2Seq learning (see Appendix \ref{app:vs} for more details). 
	Over unlabled dialog data, SEDST is trained through posterior regularization (PR), where a posterior network is used to model the posterior belief distribution given system responses, and then guide the learning of prior belief tracker through minimizing the KL divergence between them.
	Second, based on the \modelname{}-S2S model, we compare our variational learning (VL) method to a classic semi-supervised learning baseline, self-training (ST), which performs as its name suggests. Specifically, after supervised pretraining over small-sized labeled dialogs, we run the system to generate pseudo belief states $b_t$ over unlabeled dialogs, and then train the response decoder $p_\theta(r_t|b_t,c_t,d_t)$ in a supervised manner. The gradients will propagate through the discrete belief states by the Straight Through gradient estimator \cite{bengio2013estimating} over the computational graph, thus also adjusting the belief state decoder $p_\theta(b_{t}|b_{t-1}, c_t)$.
	
	
	\section{Results and Analysis}
	
	\begin{table*}[t]
		\resizebox{.99\linewidth}{!}{ 
			\begin{tabular}{cl cccc c ccc}
				\toprule
				\multirow{2}{*}{Type} & \multicolumn{1}{c}{\multirow{2}{*}{Model}} & \multicolumn{4}{c}{CamRest676}                      &~~& \multicolumn{3}{c}{In-Car}                     \\
				\cmidrule(lr){3-6} \cmidrule(lr){8-10} 
				& \multicolumn{1}{c}{}                       & Joint Goal    & Match         & SuccF1    & BLEU          &~~& Match         & SuccF1    & BLEU          \\
				\midrule
				\multirow{2}{*}{DST}  & StateNet \cite{ren2018towards}                                  & 88.9          & -             & -             & -             &~~& -             & -             & -             \\
				& TripPy \cite{heck2020trippy}                                  & 92.7$\pm$0.2      & -             & -             & -             &~~& -             & -             & -             \\
				\midrule
				\multirow{7}{*}{E2E}  & LIDM \cite{wen2017latent}                                  & 84.2$^*$          & 91.2          & 84.0          & 24.6          &~~& 72.1          & 76.2          & 17.3          \\
				& TSCP \cite{lei2018sequicity}                                       & 87.4$^*$          & 92.7          & 85.4          & 25.3          &~~& 84.5          & 81.1          & 21.9          \\
				& SEDST \cite{sedst}                                 & 88.1$^*$          & 92.7          & 75.4 & 23.6          &~~& 84.5          & \textbf{82.9} & 19.3          \\
				& FSDM \cite{fsdm}                                  & -             & 93.5          & \textbf{86.2} & 25.8 &~~& 84.8          & 82.1          & 21.5          \\
				& MOSS \cite{liang2020moss}                                   & 88.4$^*$          & 95.1          & 86.0          & \textbf{25.9} &~~& -             & -    & -             \\
				& \modelname{}-S2S (best)                                &\textbf{93.5~~} & \textbf{96.4} & 82.3          & 25.6          &~~& \textbf{86.6} & 78.0          & \textbf{23.2} \\
				& \modelname{}-S2S (statistical)                                 & 91.7$\pm$1.5      & 96.4$\pm$0.5      & 83.0$\pm$1.0      & 25.5$\pm$0.4      &~~& 85.8$\pm$1.7      & 77.0$\pm$1.7      & 22.8$\pm$1.1   \\
				\bottomrule  
		\end{tabular}}
		\caption{Results on CamRest676 and In-Car. The model with the highest joint goal accuracy on the development set of CamRest676 is shown as the best result, as similarly reported in prior work. Statistical results are reported as the mean and standard deviation of 5 runs. $^*$ denotes results obtained by our run of the open-source code. }
		\label{res_small} 
	\end{table*}

	\begin{table*}[t]
		\resizebox{.99\linewidth}{!}{ 
			\begin{tabular}{clcc c cccc}
				\toprule
				\multicolumn{4}{c}{Model Configure}                                      & \multicolumn{1}{l}{Belief Tracking} & \multicolumn{4}{c}{Response Generation}   \\
				\cmidrule(lr){1-4} \cmidrule(lr){5-5} \cmidrule(lr){6-9}
				Type                 & \multicolumn{1}{c}{Model} & Size  & Pretrained LM   & Joint Goal                          & Inform & Success & BLEU  & Combined \\
				\midrule
				\multirow{3}{*}{DST} & TRADE \cite{wu2019transferable}   & 10.2M & no              & 45.60                               & -      & -       & -     & -              \\
				& NADST \cite{le2020nadst}          & 12.9M & no              & 49.04                               & -      & -       & -     & -              \\
				& CSFN-DST \cite{zhu2020efficient}  & 63M   & no              & 50.81                               & -      & -       & -     & -              \\
				\cmidrule(lr){1-2}
				\multirow{6}{*}{E2E}& TSCP \cite{lei2018sequicity}        & 1.4M  & no              & 37.53                               & 66.41  & 45.32   & 15.54 & 71.41          \\
				& SFN + RL \cite{mehri2019structured} & 1.4M  & no              & 21.17$^*$                               & 73.80  & 58.60   & 16.88 & 83.04          \\
				& DAMD \cite{zhang2020task}           & 2.0M  & no              & 35.40$^*$                               & 76.40  & 60.40   & 16.60 & 85.00          \\
				& UniConv \cite{le2020uniconv}        & 16M   & no     & 50.14                               & 72.60  & 62.90   & \textbf{19.80} & 87.55          \\
				& \modelname{}-S2S (best)           & 3.8M  & no              &\textbf{51.45}                            &\textbf{78.07}     &\textbf{67.06}      &18.13   &\textbf{90.69}                \\
				& \modelname{}-S2S (statistical)    & 3.8M  & no              &50.05                                  &76.89 &63.30   &17.92       &88.01\\
				\midrule
				\multirow{3}{*}{DST} & CSFN-DST + BERT \cite{zhu2020efficient} & 115M  & BERT            & 52.88                               & -      & -       & -     & -              \\
				& DST-Picklist \cite{zhang2019find} & 220M  & BERT            & 53.30                               & -      & -       & -     & -              \\
				& TripPy \cite{heck2020trippy}      & 110M  & BERT            & 55.29                               & -      & -       & -     & -              \\
				\cmidrule(lr){1-2}
				\multirow{2}{*}{E2E} & SimpleTOD \cite{hosseini2020simple} & 81M   & DistilGPT-2 & 56.45                               & 85.00  & 70.05   & 15.23 & 92.98          \\
				& SOLOLIST \cite{peng2020soloist}     & 117M  & GPT-2           & -                                   & 85.50  & 72.90   & 16.54 & 95.74          \\
				\bottomrule  
		\end{tabular}}
		\caption{Results on MultiWOZ 2.1. The model with the highest validation joint goal accuracy is shown as the best result, as similarly reported in prior work. The standard deviations for the statistical results are in Table \ref{dev} in the appendix. $^*$ denotes results obtained by our run of the open-source code. }
		\label{multiwoz} 
	\end{table*}
	
	\begin{table*}[t]
		\resizebox{.99\linewidth}{!}{ 
			\begin{tabular}{c l  cccc c cccc}
				\toprule
				\multirow{2}{*}{\shortstack{Labeled\\ Data}}        & \multicolumn{1}{c}{\multirow{2}{*}{Model \& Method}} & \multicolumn{4}{c}{CamRest676}                                  &~~& \multicolumn{4}{c}{In-Car}                                   \\ \cmidrule(lr){3-6} \cmidrule(lr){8-11}  
				& \multicolumn{1}{c}{}                         & Joint Goal    & Match         & SuccF1    & \multicolumn{1}{c}{BLEU}          &~~& Joint Goal    & Match         & SuccF1    &  BLEU         \\ 
				\midrule 
				\multirow{5}{*}{50\%}& \modelname{}-S2S + SupOnly                            & 83.3          & 91.8          & 80.5          & 23.8          &~~& 77.9          & 81.0          & 74.5          & 20.4          \\ 
				
				& \modelname{}-S2S + Semi-ST                          & 86.3          & 93.1          & 83.1          & 25.3          &~~& 79.8          & 83.4          & 74.8          & 22.1          \\ 
				
				& \modelname{}-S2S + Semi-VL                           & {89.7} & {94.4} & {83.1} & {25.3} &~~& {81.1} & {84.1} & {77.5} & {22.6} \\ \cmidrule{2-11}
				
				& SEDST + SupOnly                            & 78.5          & 89.1          & 65.0          & 18.6          &~~& 74.4          & 74.1          & 69.2          & 16.9          \\ 
				& SEDST + Semi-PR                          & {79.5}          & {91.1}          & {71.2}          & {21.4}          &~~& {77.2}          & {77.8}          & {75.0}          & {19.4}          \\ 
				\midrule
				\multirow{5}{*}{25\%} & \modelname{}-S2S + SupOnly                          & 68.8          & 85.9          & 75.3          & 21.7          &~~& 74.3          & 73.7          & 62.8          & 15.8          \\ 
				& \modelname{}-S2S + Semi-ST                              & 74.1          & 91.1          & {82.5} & 25.4          &~~& 74.9          & 74.4          & {76.9} & {22.5} \\ 
				& \modelname{}-S2S + Semi-VL                               & {77.5} & {93.6} & 81.4          & {25.5} &~~& {78.8} & {79.3} & 76.6          & 22.4          \\ 
				\cmidrule{2-11}
				& SEDST + SupOnly                            & 64.2          & 80.3          & 66.8          & 16.9          &~~& 57.8          & 51.0          & 50.4          & 14.1          \\ 
				& SEDST + Semi-PR                        & {65.1}         & {83.0}          & {71.7}          & {22.1}          &~~& {63.6}          & {59.9}          & {70.4}          & {19.3}          \\ 
				\bottomrule 
		\end{tabular}}
		\caption{\textit{SupOnly} denotes training with only labeled data, and \textit{Semi} denotes training with both labeled and unlabeled data in each dataset. ST, VL and PR denote self-training, variational learning and posterior regularization \cite{sedst} respectively. Results of SEDST are obtained by our run of the open-source code. All the scores in this table are the mean from 5 runs. }
		\label{semi_small} 
	\end{table*}
	
	In our experiments, we report both the best result and the statistical result obtained from multiple independent runs with different random seeds. Details are described in the caption of each table. The implementation details of our model is available in Appendix \ref{app:imlp}. Results are organized to show the advantage of our proposed \modelname{}-S2S model over existing models (\S\ref{sec:res_sup}) and the effectiveness of our semi-supervised learning method (\S\ref{sec:semi_exp}). 
	
	\subsection{Benchmark Performance}
	\label{sec:res_sup}
	We first train our \modelname{}-S2S model under full supervision and compare with other baseline models on the benchmarks. The results are given in Table \ref{res_small} and Table \ref{multiwoz}.

	As shown in Table \ref{res_small}, \modelname{}-S2S obtains new SOTA joint goal accuracy on CamRest676 and the highest match scores on both CamRest676 and In-Car datasets. Its BLEU scores are also beyond or close to the previous SOTA models. 
	The relatively low SuccF1 is due to that in \modelname{}-S2S, we do not apply additional dialog act modeling and reinforcement fine-tuning to encourage slot token generation as in other E2E models.
	
	Table \ref{multiwoz} shows the MultiWOZ results. Among all the models without using large pretrained LMs, \modelname{}-S2S performs the best in belief tracking joint goal accuracy and 3 out of the 4 response generation metrics. Although the response generation performance is not as good as recent GPT-2 based SimpleTOD and SOLOLIST, our model is much smaller and thus computational cheaper.
	
	\subsection{Semi-Supervised Experiments}
	\label{sec:semi_exp}
\begin{figure}[t]
		\centering
		\subfigure[Joint Goal Accuracy]
		{	\label{fig_jg}
			\includegraphics[width=0.46\columnwidth]{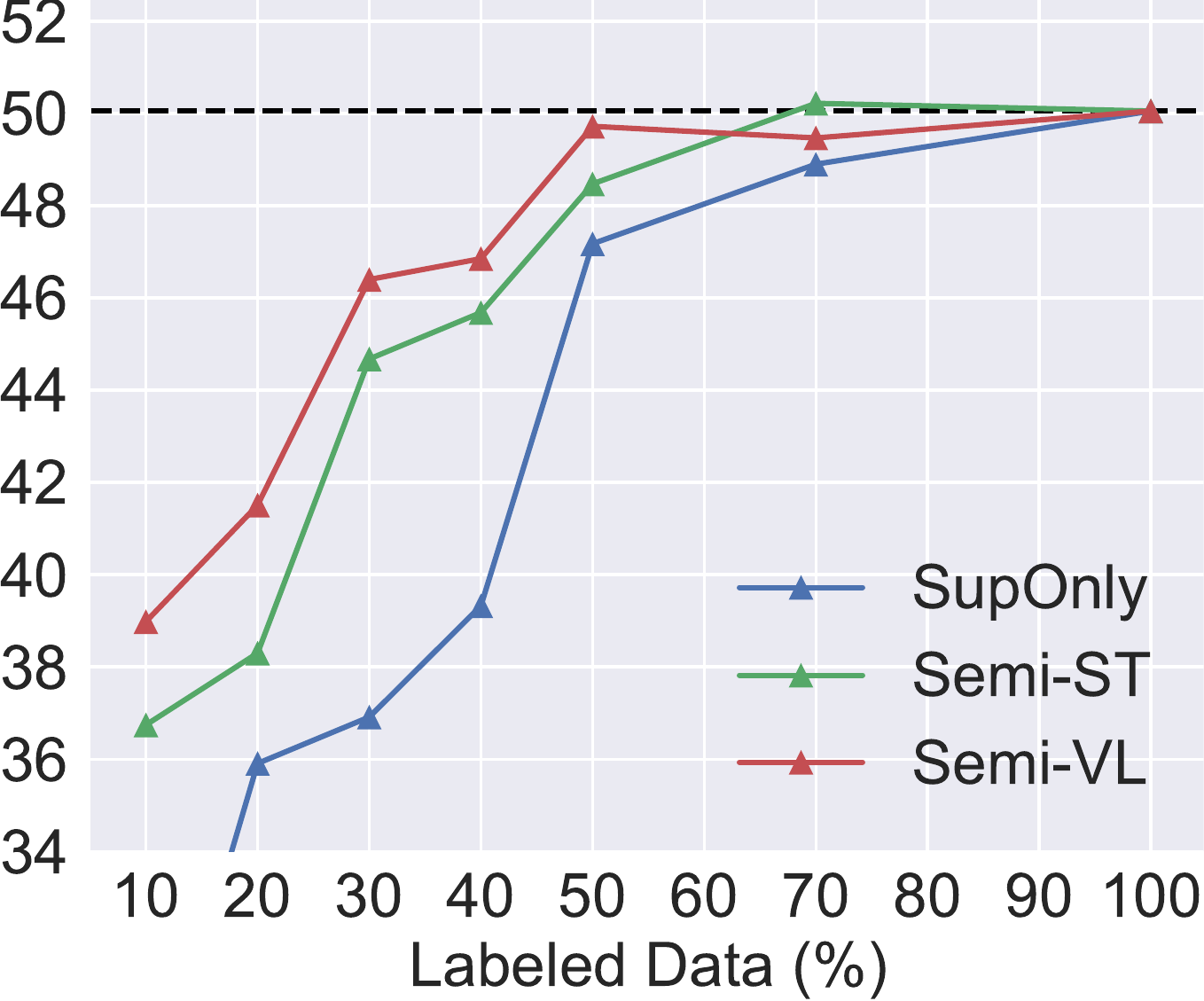} }
		\subfigure[Combined Score]
		{	\label{fig_cs}
			\includegraphics[width=0.46\columnwidth]{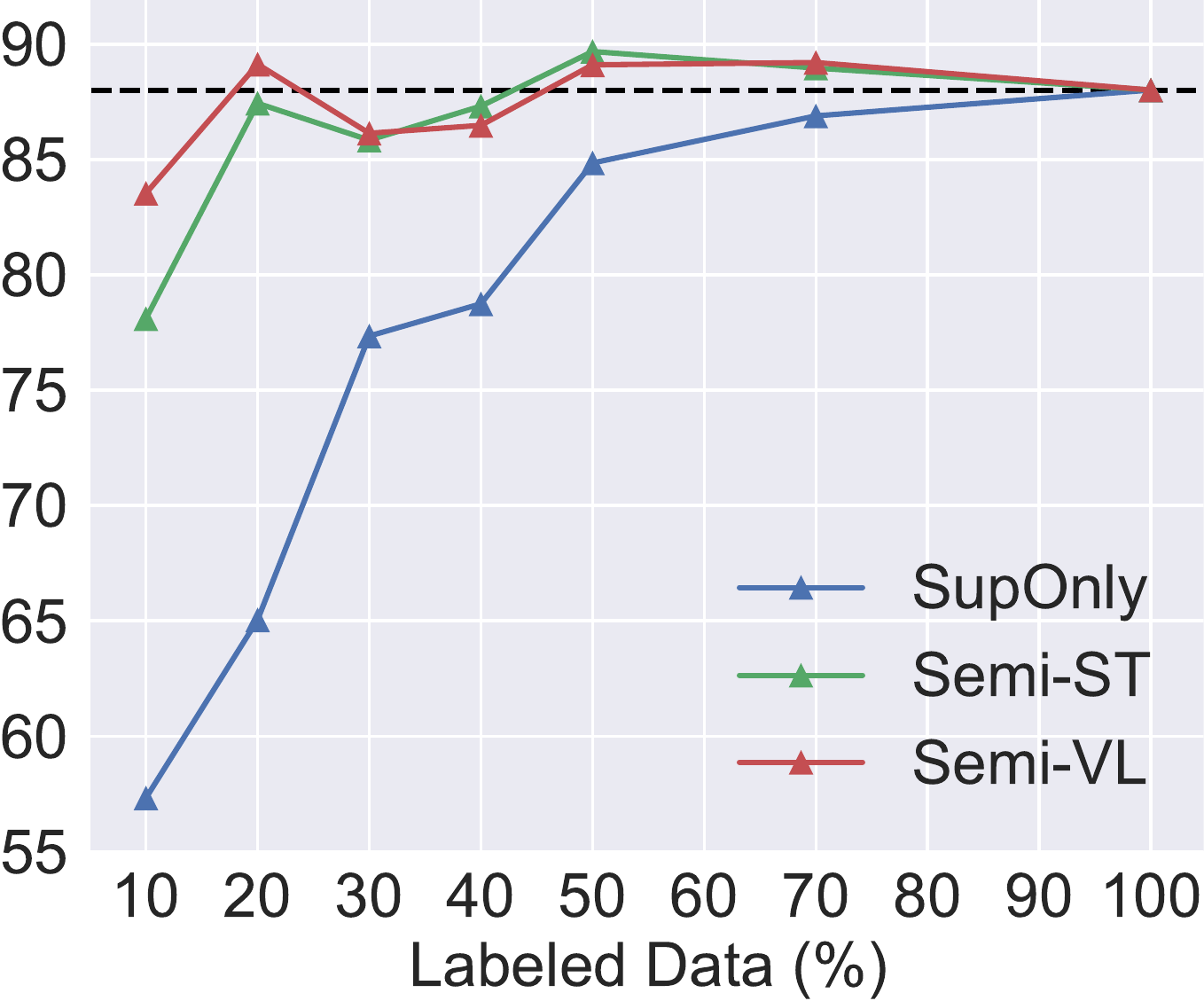} }
		\caption{Performance of different methods w.r.t labeling proportion on MultiWOZ 2.1. The dash line corresponds to the baseline trained with 100\% labeled data. }
		\label{semi_big}
	\end{figure}

	In our semi-supervised experiments, we first split the data according to a fixed proportion, then train the models using only labeled data (SupOnly), or using both labeled and unlabeled data (Semi) with the proposed variational learning method (Semi-VL), self-training (Semi-ST) and posterior regularization (Semi-PR) introduced in \S\ref{sec:baselines} respectively. We conduct experiments with 50\% and 25\% labeled data on CamRest676 and In-Car following \citet{sedst}, and change the labeled data proportion from 10\% to 100\% on MultiWOZ. The results are shown in Table \ref{semi_small} and Figure \ref{semi_big}.  
	
	In Table \ref{semi_small}, we can see that semi-supervised learning methods outperform the supervised learning baseline consistently in all experiments for the two datasets. In particular, the improvement of Semi-VL over SupOnly on our model is significantly larger than Semi-PR over SupOnly on SEDST in most metrics, and Semi-VL obtains a joint goal accuracy of 1.3\%$\sim$3.9\% higher over Semi-ST. These results indicate the superiority of our LABES modeling framework in utilizing unlabeled data over other semi-supervised baselines. Since LABES mainly improves modeling of belief states, it is more relevant to examine the belief tracking metrics such as joint goal accuracy and match rate (partly determined by the belief tracking accuracy). 
	Note that Semi-VL and Semi-ST are fed with the same set of system responses, thus they obtain similar SuccF1 and BLEU scores in Table \ref{semi_small}, which mainly measure the response quality.
	
	\begin{table}[t!]
		\centering
		\resizebox{.99\columnwidth}{!}{ 
			\begin{tabular}{cl}
				\toprule
				\multicolumn{2}{c}{Dialog \#586 in CamRest676 } \\
				\midrule
				\multicolumn{2}{l}{$u_1$: I am looking for an expensive restaurant that serves Russian food.} \\
				\multicolumn{2}{l}{$b_1$: $\{$food: Russian, pricerange: expensive$\}$} \\
				\multicolumn{2}{l}{$r_1$: There is no expensive restaurant that serves Russian food. Can I } \\
				\multicolumn{2}{l}{~~~~~  help you with anything else?} \\
				\specialrule{0em}{2pt}{2pt}
				\multicolumn{2}{l}{$u_2$: Yes, do you have British type food ?}  \\
				\multicolumn{2}{l}{$b_2$: $\{$food: British, pricerange: expensive$\}$}  \\
				\multicolumn{2}{l}{$r_2$: Yes, there are 6 options. Does the part of town matter?} \\
				\specialrule{0em}{2pt}{2pt}
				\multicolumn{2}{l}{$u_3$: I don't care. }\\
				\midrule
				\multirow{2}{*}{\shortstack{Sup\\Only}} & $b_3$: $\{$food: British, pricerange: expensive$\}$\\
				& $r_3$: $[$v.name$]$ is located at $[$v.addr$]$. How else I can help you with? \\
				\midrule
				\multirow{3}{*}{\shortstack{Semi\\-VL}} & $b_3$: $\{$food: British, pricerange: expensive, area: dontcare$\}$ \\
				& $r_3$: $[$v.name$]$ is a nice British restaurant in the $[$v.area$]$ of town. \\
				& ~~~~~ Would you like their address? \\
				\midrule
				\multicolumn{2}{c}{Dialog \#PMUL1329 in MultiWOZ } \\
				\midrule
				\multicolumn{2}{l}{$u_1$:  I need a train going to Bishops Stortford.} \\
				\multicolumn{2}{l}{$b_1$: $\{$destination: Bishops Stortford$\}$} \\
				\multicolumn{2}{l}{$r_1$: OK. Are you departing from Cambridge?} \\
				\specialrule{0em}{2pt}{2pt}
				\multicolumn{2}{l}{$u_2$: Yes, that's why I'm requesting help from Cambridge towninfo.}\\
				\midrule
				\multirow{2}{*}{\shortstack{Sup\\Only}} & $b_3$: $\{$destination: Bishops Stortford$\}$\\
				& $r_3$: Where will you be departing from? \\
				\midrule
				\multirow{3}{*}{\shortstack{Semi\\-VL}} & $b_3$: $\{$destination: Bishops Stortford, departure: Cambridge$\}$ \\
				& $r_3$: I have $[$v.choice$]$ trains that meet your criteria. What day \\
				& ~~~~~ would you like to travel? \\
				\bottomrule
		\end{tabular}}
		\caption{Comparison of two example turns generated by our model with supervised learning only (SupOnly) and semi-supervised variational learning (Semi-VL). }
		\label{case} 
	\end{table}
	
	The results on MultiWOZ shown in Figure \ref{semi_big} also support the above conclusions. From the plot of metric scores w.r.t labeling  proportions, we can see how many labels can be reduced clearly. Our \modelname{}-S2S model trained with Semi-VL obtains a joint goal accuracy of 49.47\% and a combined score of 89.21 on only 50\% of labeled data, which is very close to 50.05\% and 88.01 obtained under 100\% supervision. This indicates that we can reduce 50\% of labels without losing performance, which results in reducing around 30,000 belief state annotations given the size of MultiWOZ.
	Moreover, it can be seen from Figure \ref{semi_big} that our Semi-VI can improve the belief tracking and response generation performance when labeling only 10\% of dialogues, and the smaller amount of labels, the larger gain obtained by Semi-VI. 
	
	\subsection{Case Study}
	We give two examples where the model trained with Semi-VL improves over the supervised-training-only baseline. In both examples, the user indicates his/her goal implicitly with a short reply. These rarely occurred corner cases are missed by the baseline model, but successfully captured after semi-supervised learning. Moreover, we can see that Semi-VL helps our model learn the cue word ``British" which contributes to a more informative response in the first dialog, and in the second dialog, avoid the incoherent error caused by error propagation, thus improve the response generation quality.

	\section{Conclusion and Future Work}
	In this paper we are interested in reducing belief state annotation cost for building E2E task-oriented dialog systems. We propose a conditional generative model of dialogs - \modelname{}, where belief states are modeled as latent variables, and unlabeled dialog data can be effectively leveraged to improve belief tracking through semi-supervised variational learning. 
	Furthermore, we develop LABES-S2S, which is a copy-augmented Seq2Seq model instantiation of LABES.
	We show the strong benchmark performance of \modelname{}-S2S and the effectiveness of our semi-supervised learning method on three benchmark datasets. In our experiments on MultiWOZ, we can save around 50\%, i.e. around 30,000 belief state annotations without performance loss. 
	
	There are some interesting directions for future work.
	First, the \modelname{} model is general and can be enhanced by, e.g. incorporating large-scale pre-trained language models, allowing other options for the belief state decoder and the response decoder such as Transformer based.
	Second, we can analogously introduce dialog acts $a_{1:T}$ as latent variables to define the joint distribution $p_\theta(b_{1:T}, a_{1:T}, r_{1:T}|u_{1:T})$, which can be trained with semi-supervised learning and reinforcement learning as well.
	
	\section*{Acknowledgments}
	This work is supported by NSFC 61976122, Ministry of Education and China Mobile joint funding MCM20170301.
	
	\bibliographystyle{acl_natbib}
	\bibliography{emnlp2020}
	
	\appendix
	
	\section{Model Comparisons with Prior Work}
	\label{app:vs}
\begin{figure}[t]
		\centering
		\includegraphics[width=1.0\columnwidth]{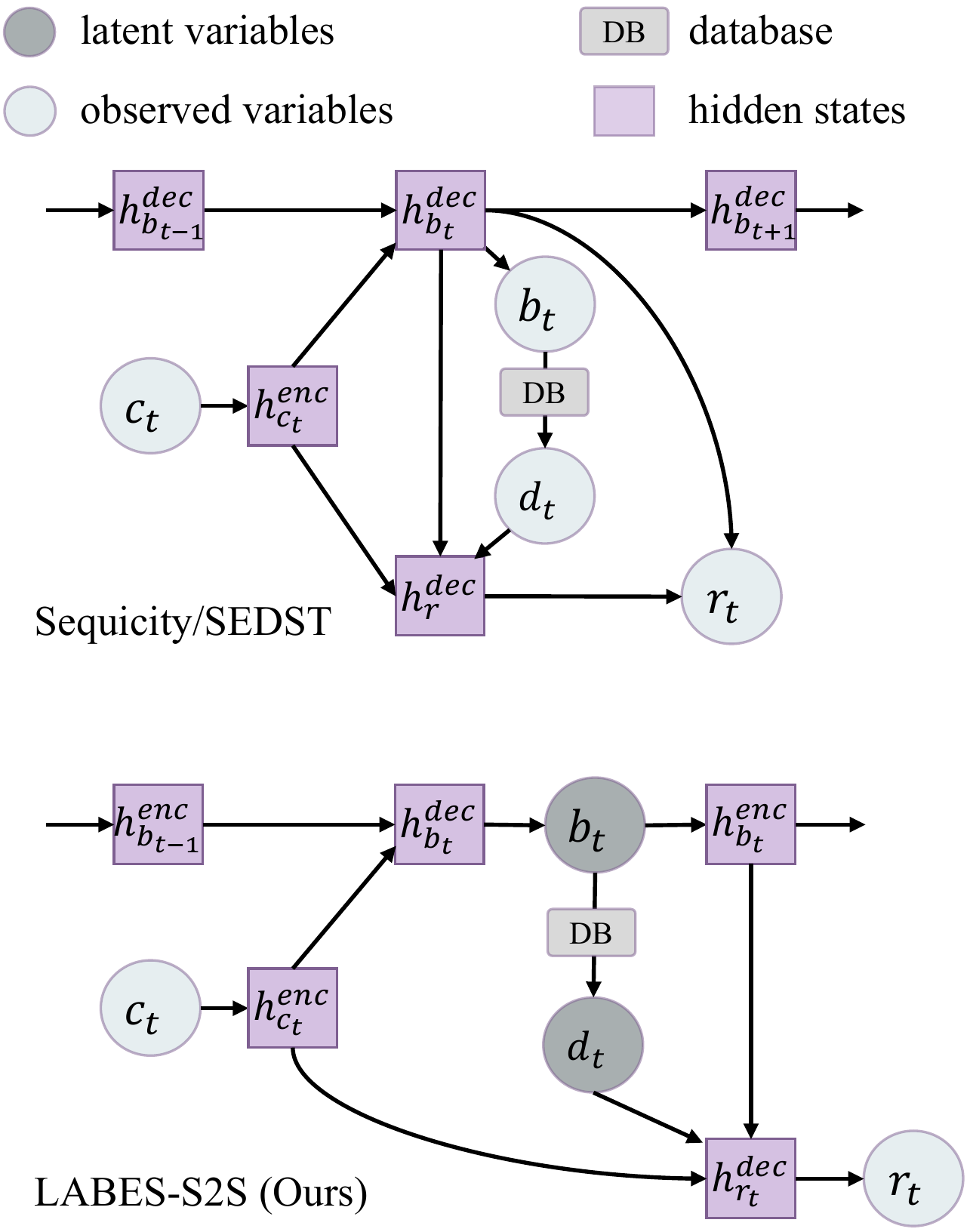}
		\caption{Comparison of computational graphs.  }
		\label{model_vs}
	\end{figure}

\begin{figure}[t]
		\centering
		\includegraphics[width=1.0\columnwidth]{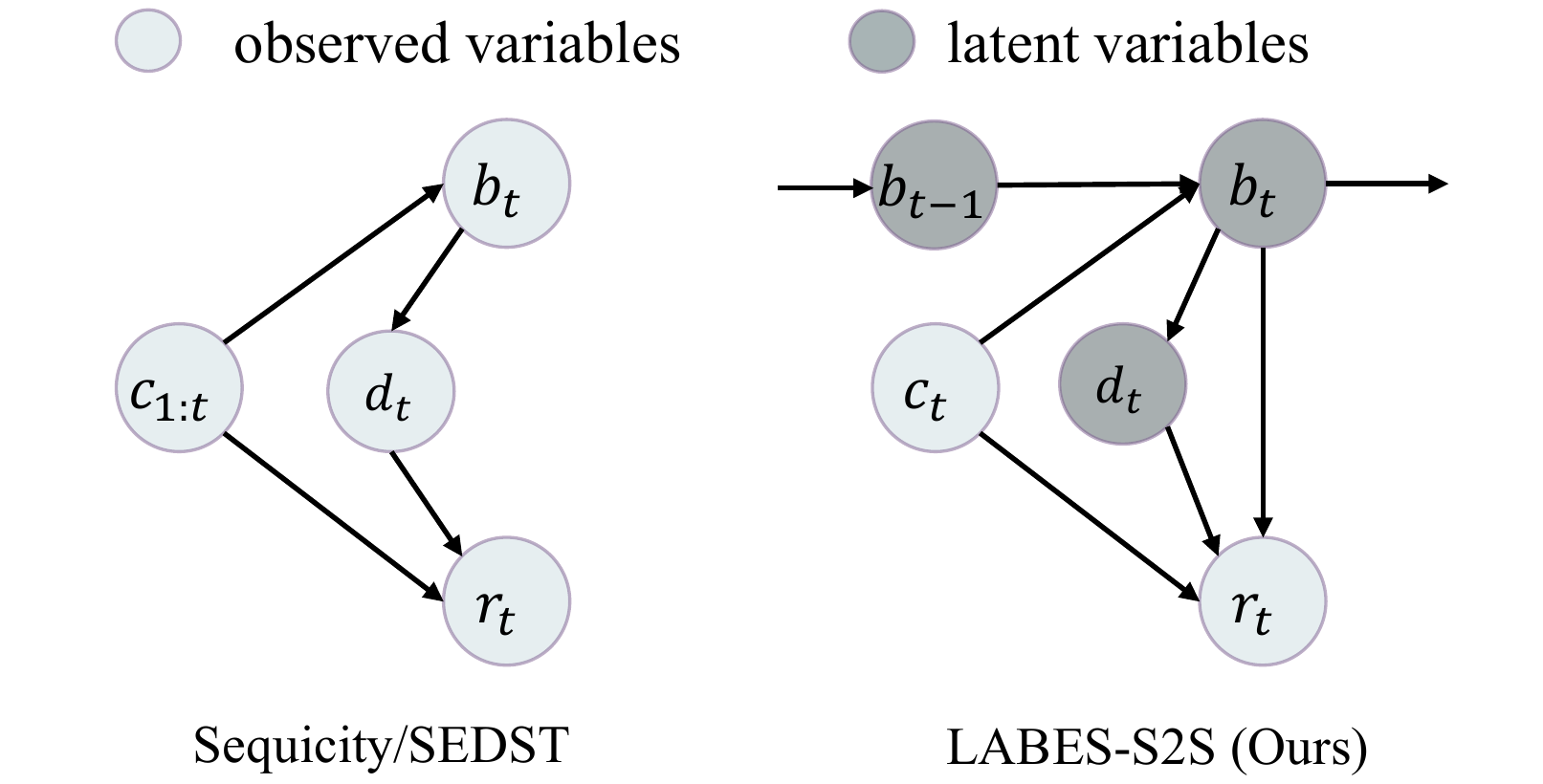}
		\caption{Comparison of probabilistic graphical model structures. }
		\label{pgm_vs}
	\end{figure}

	In this section, we comment on the differences between our \modelname{}-S2S model and Sequicity \cite{lei2018sequicity} in both models and learning methods. Note that SEDST \cite{sedst} employs the same model structure as Sequicity. 
	First, Figure \ref{model_vs} shows the difference in computational graphs between Sequicity/SEDST and \modelname{}-S2S.
	For Sequicity/SEDST, $b_t$ and $r_t$ are decoded directly from the belief state decoder's hidden states $h_{b_t}^{dec}$, thus the conditional probability of $r_t$ given $b_t$ and the state transition probability between $b_{t-1}$ and $b_t$ are not considered\footnote{Strictly speaking, the transition between belief states across turns and the dependency between system responses and belief states are modeled very weakly in Sequicity/SEDST, only owing to the copy mechanism. For simpliciy, we omit such relations in both Figure \ref{model_vs} and \ref{pgm_vs}.}. 
	In contrast, \modelname{}-S2S model introduces an additional $b_t$ encoder and uses the encoder hidden states $h_{b_t}^{enc}$ to generate system response and next turn's belief state, thus the conditional probability $p_\theta(r_t|b_t,c_t)$ and state transition probability $p_\theta(b_{t}|b_{t-1},c_t)$ are well defined by two complete Seq2Seq processes. 
	
	Second, the difference in models can also be clearly seen from the probabilistic graphical model structures as shown in Figure \ref{pgm_vs}. 
	\modelname{}-S2S is a conditional generative model where the belief states are latent variables. In contrast, Sequicity/SEDST do not treat the belief states as latent variables. 
	
	Third, the above differences in models lead to differences in learning methods for Sequicity/SEDST and \modelname{}-S2S.
	Sequicity can only be trained on labeled data via multi-task supervised learning.
	SEDST resorts to an ad-hoc combination of posterior regularization and auto-encoding for semi-supervised learning.
	Remarkably, \modelname{}-S2S is optimized under the principled variational learning framework.
	
	\begin{table*}[t]
		\resizebox{.99\linewidth}{!}{ 
			\begin{tabular}{c c cccc}
				\toprule
				\multirow{2}{*}{Model}  & \multicolumn{1}{l}{Belief Tracking} & \multicolumn{4}{c}{Response Generation}   \\
				\cmidrule(lr){2-2} \cmidrule(lr){3-6}
				& Joint Goal                          & Inform & Success & BLEU  & Combined \\
				\midrule
				\modelname{}-S2S (statistical)            &50.05${\pm0.92}$                                  &76.89${\pm1.51}$ &63.30${\pm2.35}$   &17.92${\pm0.35}$       &88.01${\pm2.10}$\\
				\bottomrule  
		\end{tabular}}
		\caption{Statistical results of our LABES-S2S model with standard deviations on MultiWOZ 2.1. }
		\label{dev} 
	\end{table*}
	
	\section{Datasets}
	\label{app:data}
	In our experiments, we evaluate different models on three benchmark task-oriented datasets with different scales and ontology complexities (Table \ref{dataset}). The Cambridge Restaurant (CamRest676) dataset \cite{wen2017a} contains single-domain dialogs where the system assists users to find a restaurant. The Stanford In-Car Assistant (In-Car) dataset \cite{eric2017key} consists of dialogs between a user and a in-car assistant system covering three tasks: calendar scheduling, weather information retrieval and point-of-interest navigation. The MultiWOZ \cite{budzianowski2018multiwoz} dataset is a large-scale human-human multi-domain dataset containing dialogs in seven domains including attraction, hotel, hospital, police, restaurant, train, and taxi. It is more challenging due to its multi-domain setting, complex ontology and diverse language styles. As there are some belief state annotation errors in MultiWOZ, we use the corrected version MultiWOZ 2.1 \cite{eric2019multiwoz} in our experiments. We follow the data preprocessing setting in \citet{zhang2020task}, whose data
	cleaning is developed based on \citet{wu2019transferable}.
	
	\begin{table}[h]
		\resizebox{.99\linewidth}{!}{
			\begin{tabular}{lccc}
				\toprule
				& CamRest676 & In-Car & MultiWOZ       \\
				\midrule
				\#Dialog     & 676       & 3031              & 10438 \\ 
				Avg. \#Turn  & 4.1                & 5.2                       & 6.9            \\ 
				\#Domain     & 1                  & 3                         & 7              \\ 
				\#Info. Slot & 3                  & 11                        & 31             \\
				\#Req. Slot  & 7                  & 11                        & 38             \\
				\#Values     & 99                 & 284                       & 4510           \\ \bottomrule
		\end{tabular}}
		\caption{Statistics of dialog datasets. Info and Req are shorthands for informable and requestable respectively. }
		\label{dataset} 
	\end{table}

	\section{Implementation Details}
	\label{app:imlp}
	In our implementation of \modelname{}-S2S, we use 1-layer bi-directinonal GRU as encoders and standard GRU as decoders. The hidden sizes are 100/100/200, vocabulary sizes are 800/1400/3000, and learning rates of Adam optimizer are $3\mathrm{e}^{-3}$/$3\mathrm{e}^{-3}$/$5\mathrm{e}^{-5}$ for CamRest676/In-Car/MultiWOZ respectively. In all experiments, the embedding size is 50 and we use GloVe \cite{pennington2014glove} to initialize the embedding matrix. Dropout rate is 0.35 and $\lambda$ for variational inferece is 0.5, which are selected via grid search from $\{0.1,0.15,0.2,0.25,0.3,0.35,0.4,0.45,0.5\}$ and $\{0.1,0.3,0.5,0.7,1.0,1.5\}$ respectively. The learning rate decays by half every 2 epochs if no improvement is observed on development set. Training early stops when no improvement is observed on development set for 4 epochs. We use 10-width beam search for CamRest676 and greedy decoding for other datasets. All the models are trained on a NVIDIA Tesla-P100 GPU.



\end{document}